\documentclass{article} 
\usepackage[preprint]{colm2026_conference}

\usepackage{microtype}
\usepackage{hyperref}
\usepackage{url}
\usepackage{booktabs}
\usepackage{natbib}

\usepackage{graphicx}
\usepackage{subcaption}

\usepackage{tabularx}   
\usepackage{caption}    
\usepackage{float}
 
\usepackage{lineno}

\definecolor{darkblue}{rgb}{0.01, 0.05, 0.8}
\hypersetup{colorlinks=true, citecolor=darkblue, linkcolor=darkblue, urlcolor=darkblue}

\title{Accuracy and Reliability of Large Language Models in Cosmetic Chemistry and Skin Health: A Benchmarking Study}

\author{Amelia Liu\thanks{ 
All Python code used for this project is publicly available at the
 paper's GitHub repository \url{https://github.com/ameliaxl}} \\
Independent Researcher\\
Basking Ridge, NJ, 07920, USA \\
\texttt{amelialiu2027@gmail.com} 
}

\begin{document}

\ifcolmsubmission
\linenumbers
\fi

\maketitle

\begin{abstract}
As consumers increasingly turn to AI chatbots for skincare advice, the technical accuracy of Large Language Models (LLMs) in cosmetic chemistry remains largely under-evaluated. We benchmarked 14 LLMs on a structured set of topics related to cosmetic chemistry, including the chemical properties of specific cosmetic ingredients and common cosmetic scenarios that may be of interest to consumers. Web search was disabled throughout to assess each model's internalized knowledge rather than its internet retrieval capacity. Overall performance was poor, with the most pronounced deficits in quantitative reasoning and structural identification tasks. While models handled general skincare questions with reasonability, responses consistently lacked the technical depth required for informed consumer decision-making. Notably, conversation with AI can pose a risk: outputs that sound authoritative but contain technical errors are less likely to generate skepticism compared to responses that explicitly acknowledge uncertainty. These findings suggest that general-purpose LLMs, trained predominantly on unverified public data, are currently not reliable sources of cosmetic chemistry information. Progress on two fronts, fine-tuning verified chemical and dermatological datasets, and substantial improvements to algorithmic reasoning, will likely be needed before these tools can be considered as resources for public use.
\end{abstract}

\section{Introduction}

In recent years, large language models (LLMs) have achieved significant global prominence by demonstrating capabilities that mirror human cognitive processing across a wide array of domains \citep{Naveed:2023aa}. Since the debut of ChatGPT, these computational tools have become integrated into the fabric of daily life, assisting millions of users with tasks ranging from academic assignments and business automation to specialized inquiries in theoretical physics and skin care guidance \citep{Hostinger:2025aa}.

The architecture of LLMs is rooted in deep learning principles designed to synthesize vast corpora of human knowledge \citep{Naveed:2023aa}. These systems utilize probabilistic modeling to predict linguistic sequences and generate coherent text. While their contemporary success is often associated with the 2017 introduction of transformer models and self-attention mechanisms, their lineage extends back to the neural networks of the 1950s and the rudimentary chatbots of the 1960s \citep{vaswani2017attention, Toloka-AI:2025aa}. By early 2026, the sector had matured into a highly competitive landscape; \citet{OpenAI:2026aa} claimed that ChatGPT is a market leader with approximately 900 million weekly active users. Other influential architectures include xAI's Grok, Google's Gemini, Meta's Llama, Anthropic's Claude, DeepSeek, and Mistral AI.

Cosmetic chemistry is a specialized branch of the chemical sciences concerned with the formulation, development, and evaluation of aesthetic and personal care products \citep{Rosen:2015aa}. For consumers, cosmetics and over-the-counter skincare products serve as practical tools for hygiene and the management of dermatological concerns such as photoaging, acne, and hyperpigmentation \citep{Goh:2023aa}. As professional dermatological consultation is often cost-prohibitive and time-intensive, many consumers rely on online resources that vary considerably in quality and scientific rigor. LLMs offer a notable alternative through their immediate availability, interactive format, and authoritative tone \citep{Carrara:2026aa}.

Despite their growing presence in consumer-facing contexts, the scientific accuracy and reliability of LLM-generated responses within the specialized domain of cosmetic chemistry remain poorly characterized. Whether AI tools can provide sound skin health guidance is a matter of ongoing debate \citep{Ferreira:2023aa}. Although the conversational accessibility of these models may benefit general consumers, their application for medical or technical advice is frequently cautioned against on grounds of accuracy and safety \citep{goktas2024assessing}.

\citet{Leon:2023aa} were among the first to evaluate the use of ChatGPT in general chemistry. They found that while ChatGPT demonstrated ``conversational competence'' in general chemistry, it failed in rigorous problem-solving. \citet{Guo:2023aa} benchmarked LLMs in eight critical chemistry tasks. They found that even the most advanced LLMs exhibit inconsistent performance when moving from general explanation to technical reasoning. While these papers focused on general chemistry, we are interested in the specialized domain of cosmetic chemistry. In the field of clinical dermatology, the initial assessments of LLMs by \citet{Ferreira:2023aa} suggest a high rate of 'appropriate' responses to patient queries. However, the underlying scientific rigor regarding ingredient chemistry remains largely unquantified.

This study evaluates 14 popular LLMs from seven leading AI developers across five topic areas in cosmetic chemistry and skin health, spanning both quantitative tasks and qualitative skincare scenarios. The findings reveal a consistent performance disparity between retrieval-based tasks, where several models performed well, and those requiring multi-step chemical reasoning, where most models struggled regardless of parameter count. These results highlight both the promise and the current limitations of LLMs as informational tools at the intersection of cosmetic chemistry and dermatology.

\section{Experiment Design and Method}
\subsection{Large Language Models}

This study evaluates a diverse cohort of LLMs from seven prominent AI developers, including Anthropic, DeepSeek, Google, Meta, Mistral AI, OpenAI, and xAI, to assess their proficiency in resolving cosmetic chemistry inquiries. Fourteen models were selected for this benchmarking experiment, with two variants drawn from each company: a flagship ``Pro" or ``Large" model, typically characterized by a high parameter count, and a scaled-down ``Flash" or ``Mini" variant designed for efficiency.

The architectural capacity of these models is heavily influenced by their parameter counts, which are established during pre-training on vast heterogeneous datasets. While parameter count often serves as a proxy for raw cognitive capability, it is not the sole determinant of success. Factors such as training data curation, architectural efficiency (e.g., Mixture-of-Experts), and reinforcement learning from human feedback (RLHF) contribute significantly to a model's specialized performance.
The Chinchilla scaling laws proposed by \citet{Hoffmann:2022aa} suggests that models with fewer parameters can outperform larger models if they are trained on sufficiently large and high-quality datasets. 
A goal of this study is to examine the relationship between scale and accuracy in the specialized field of cosmetic chemistry.

The specific model architectures and their respective parameter scales are detailed in Table 1.\footnote[2]{Estimates for proprietary systems (GPT-5, Gemini 2.5 Pro/Flash, Grok 4, and the Claude 4.5 suite) are derived from consensus industry benchmarks, whereas figures for the Meta Llama 4 series, Mistral AI, and DeepSeek R1 are based on publicly disclosed technical specifications.}
The selection criteria prioritized both market dominance and geographic diversity. OpenAI's ChatGPT and Google's Gemini represent the most pervasive platforms in the consumer market. Anthropic, Meta, and xAI represent leading American research initiatives, while DeepSeek and Mistral AI provide critical perspectives from the burgeoning Chinese and European AI sectors, respectively. By spanning a wide spectrum of training methodologies and regional datasets, this multi-model approach ensures a comprehensive evaluation of current AI capabilities in cosmetic science.

Each model was subjected to a rigorous evaluation involving a predefined set of cosmetic chemistry prompts, ranging from structural bond analysis to complex group-contribution calculations.

\begin{table}[t]
    \centering
    \caption{Comparative Model Architectures and Parameter Scales of Tested LLMs (Q1 2026)}
    \label{tab:llm-models}
    \small
    \begin{tabular}{@{}llrlrc@{}}
        \toprule
        \textbf{Developer} & \textbf{Flagship Model} & \textbf{Parameters} & \textbf{Efficiency Variant} & \textbf{Parameters} \\
        \midrule
        OpenAI    & GPT-5             & $\sim$1,800 B & GPT-5 Mini          & $\sim$149 B \\
        Google    & Gemini 2.5 Pro    & $\sim$288 B & Gemini 2.5 Flash    & $\sim$17 B  \\
        Meta      & Llama 4 Maverick  & $\sim$402 B & Llama 4 Scout       & $\sim$108 B \\
        xAI       & Grok 4            & $\sim$1,700 B & Grok 4 Fast         & $\sim$314 B \\
        Anthropic & Claude 4.5 Opus   & $\sim$175 B & Claude 4.5 Haiku    & $\sim$20 B  \\
        Mistral AI & Mistral Large 3   & $\sim$675 B & Mistral 14B         & 14 B        \\
        DeepSeek  & DeepSeek R1       & $\sim$685 B & R1-Distill-Qwen-32B & 32 B        \\
        \bottomrule
        \addlinespace
        \multicolumn{5}{l}{\footnotesize \textit{Note: B = Billion. Parameter counts for proprietary models are based on industry estimates.}}
    \end{tabular}
\end{table}

\subsection{Experimental Prompts and Target Stimuli}

To evaluate the proficiency of LLMs in synthesizing both quantitative and qualitative knowledge within the cosmetic domain, each model was subjected to five distinct thematic inquiries. The prompts were structured to elicit concise, one-sentence responses, ensuring a focused assessment of factual accuracy and technical reasoning. The specific prompts are provided in Appendix \ref{appendix:prompts}.

\textbf{Quantitative Assessment: Chemical Properties and Reasoning.} The first four topics evaluated the LLMs' command of structural chemistry and computational logic. These tasks represent foundational inquiries for which responses can be independently validated:

\begin{itemize}
\item \textbf{Topic 1 --- Molecular Bonding Analysis:}
Identify the number of sigma and pi bonds for five cosmetic ingredients: Squalane, Phenoxyethanol, Sodium Lauryl Sulfate, Isododecane, and Chlorphenesin.

    \item \textbf{Topic 2 --- Compositional Analysis:} Identify 
    synthetic preservatives within five prominent formulations, including \textit{CeraVe Hydrating Toner}, \textit{Kiehl's Ultra Facial Cream}, and \textit{The Ordinary Hyaluronic Acid 2\% + B5} (2024 reformulation).

    \item \textbf{Topic 3 --- Molecular Weight Determination:} Find  the molecular weight (rounded to one decimal place) for Squalane, Octane, Dodecane, Isopropyl Alcohol, and Chlorphenesin.
    
    \item \textbf{Topic 4 --- Thermophysical Estimation (Joback Method):} Calculate normal boiling points in Kelvin using the Joback method \citep{Joback:1987aa}. This task tested multi-step algorithmic reasoning beyond simple data retrieval.
\end{itemize}

 \textbf{Qualitative Assessment of Skincare Advice:}
 \textbf{Topic 5} asked LLMs to provide consumer-oriented guidance in two real-world contexts: (1) \textbf{Sequential skincare advice}, describing subsequent steps for sensitive, acne-prone skin following specific interventions (e.g., azelaic acid, cleansing); and (2) \textbf{First-step recommendations}, identifying initial
treatments for concerns such as hyperpigmentation and excess sebum.

\textbf{Methodological Rigor and Validation.} All experimental scenarios were conceived before data collection to prevent any form of post-hoc selection. Test stimuli, including specific cosmetic ingredients and finished products, were curated based on their frequency and significance of use in personal care. Experimental scenarios were further designed to simulate high-frequency, real-world consumer interactions.  Quantitative responses were cross-referenced against NCBI \textit{PubChem} \citep{Kim:2023aa}. Qualitative outputs were benchmarked against foundational textbooks such as \textit{Harry's Cosmeticology} 
\citep{Rosen:2015aa}.

\subsection{Experimental Design}

All interactions with the selected LLMs were conducted systematically to ensure consistency and reproducibility across models. To interface with multiple LLMs  through a single, standardized endpoint, we used the OpenRouter API gateway, a unified gateway offering access to hundreds of large language models across different providers \citep{OpenRouter:2025aa}. Queries were automated using the official OpenAI Python library \citep{OpenAI:2025ab}, which enabled each predefined inquires to be submitted to all 14 models under identical conditions, with responses collected programmatically for subsequent analysis. 

\textbf{Prompt standardization and bias mitigation}: To maintain a controlled experimental environment, each inquiry was submitted as a discrete, independent request. To mitigate the influence of conversational drift or ``chain-of-thought'' biases, the system was configured to prevent context retention between queries. Each model formulated its response based solely on the immediate prompt, free from the carryover effects or external guidance that might artificially enhance performance. All prompts were delivered in a ``zero-shot'' format, without supplementary instructions or system-level hints.

\textbf{Control for hallucination and inference integrity}:
AI hallucination, the tendency of a model to generate responses that are fluent and coherent yet factually inaccurate or entirely fabricated, remains a documented challenge in the field of LLMs \citep{Huang:2024aa}. Research suggests that vague or ambiguous inputs frequently exacerbate hallucination by forcing LLMs to rely probabilisticaly on their training data rather than factual synthesis. Using clear, consistent, and unambiguous questions and eliminating 
iterative dialogue commonly observed in web-based chatbot interactions were therefore an intentional methodological safeguard against this risk.

\textbf{Assessment of intrinsic knowledge vs. web retrieval}:
A critical distinction was made between a model's intrinsic parameter-based knowledge and its ability to perform real-time information retrieval. Consequently, all API interactions were configured to disable web-access functionality. This design choice ensured that responses were derived exclusively from the models' pre-trained corpora, frozen at their respective knowledge cutoff dates. This restriction served as a control against the potential for Retrieval-Augmented Generation (RAG) to obscure the true reasoning capabilities of the underlying model architectures
\citep{lewis2020rag}.

\subsection{Data Analysis}

LLM outputs were extracted programmatically using Python and organized into a summary table of accuracy percentage for each cosmetic chemistry query and for each model. Accuracy for each query was calculated as the number of correct responses divided by the total number of ingredients or products queried. To examine whether parameter of LLMs predicts accuracy, a simple linear regression was conducted using the \texttt{SciPy} library, with statistical significance defined as $p < 0.05$. For the qualitative questions, response content was analyzed using word clouds using the \texttt{WordCloud} Python library.
This is a well-established visualization technique in which word size reflects frequency of occurrence, providing an intuitive overview of the most prominent themes across model responses \citep{Heimerl:2014aa}. All Python code used in these analyses is publicly available at the corresponding author's GitHub repository.

\begin{figure}[t]
\centering

\begin{subfigure}{0.48\textwidth}
\centering
\includegraphics[height=2in]{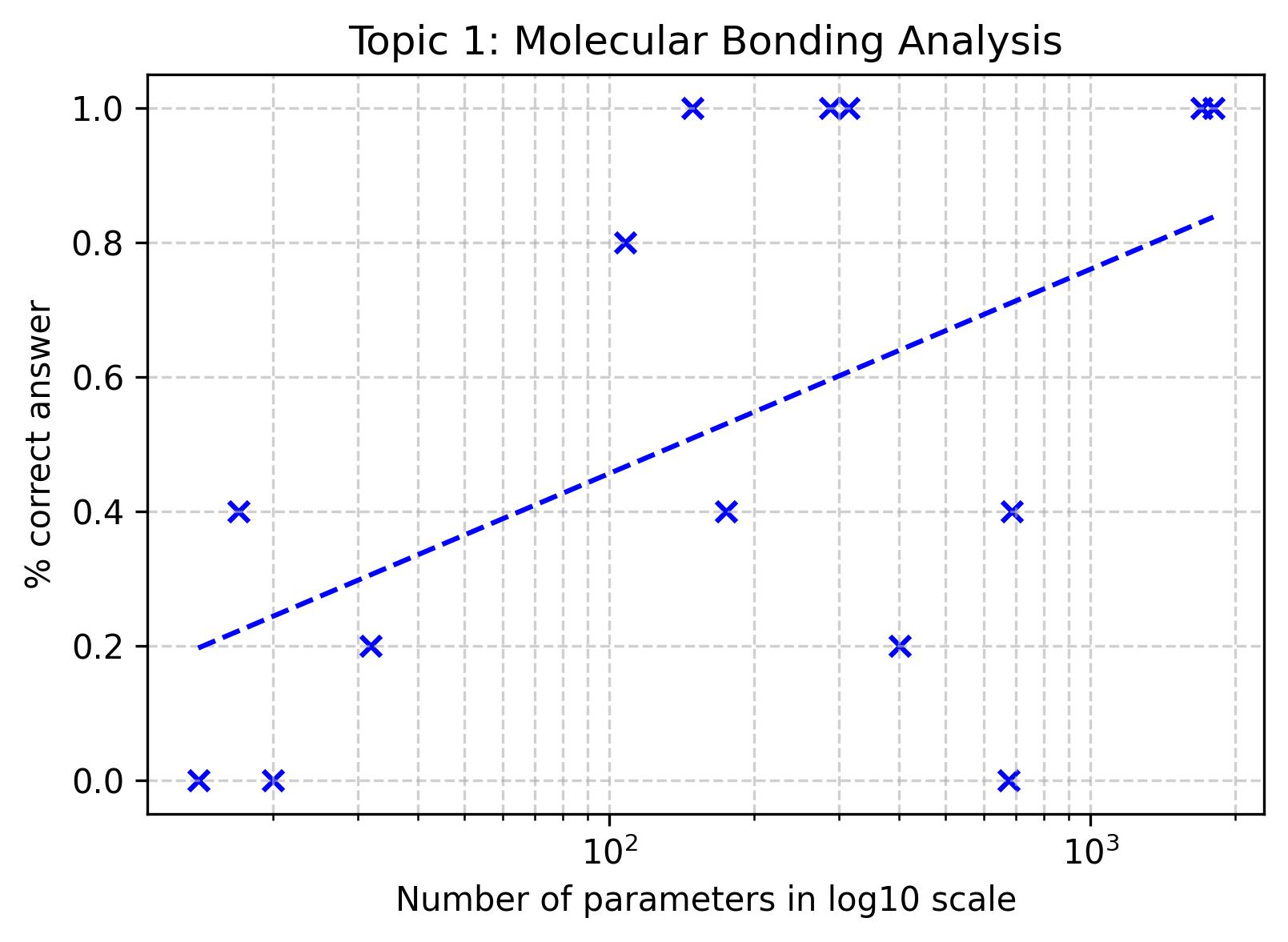}
\caption{Topic 1: Calculating Sigma and Pi Bonds}
\label{fig:1a}
\end{subfigure}
\hfill
\begin{subfigure}{0.48\textwidth}
\centering
\includegraphics[height=2in]{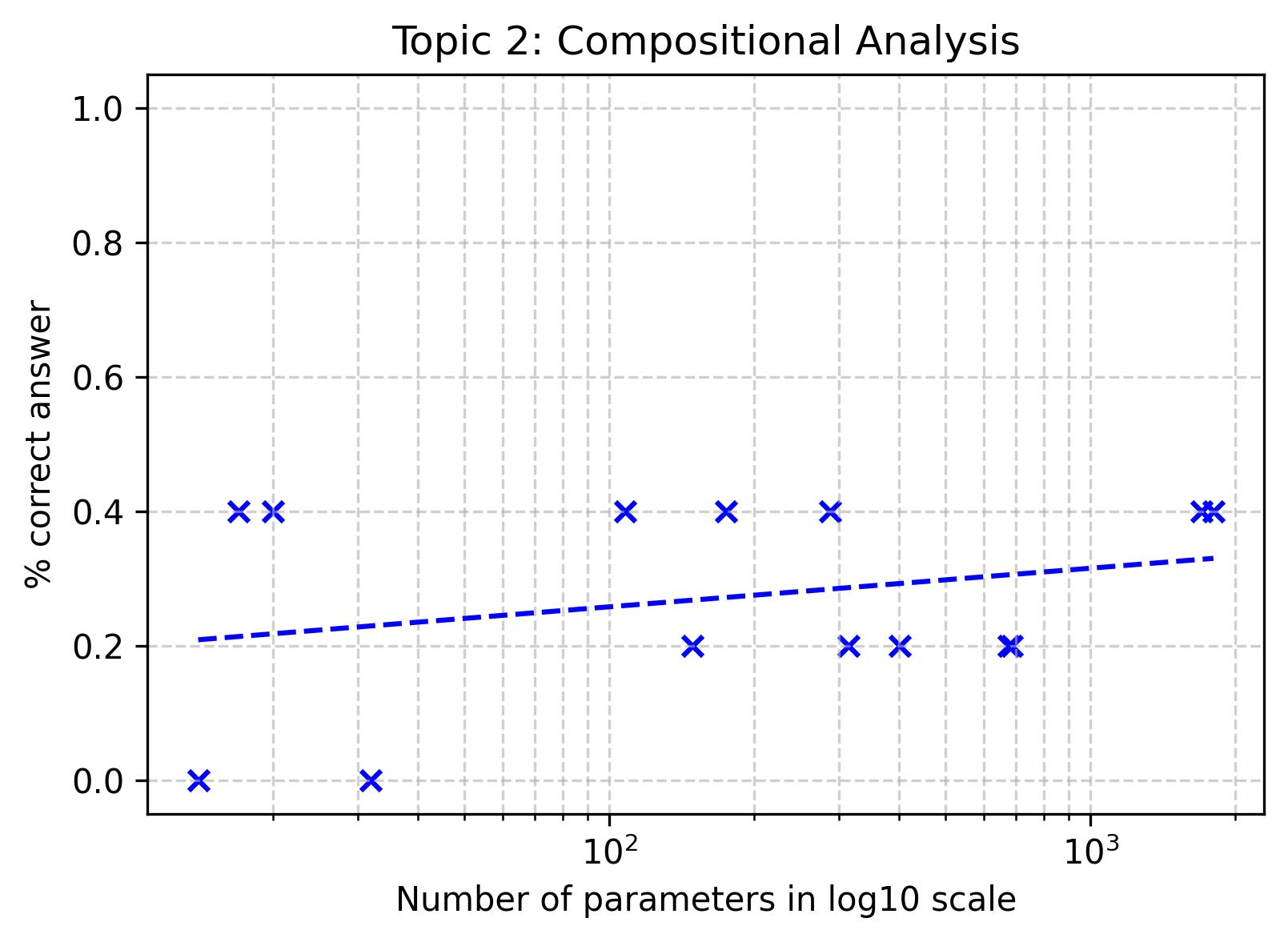}
\caption{Topic 2: Preservatives Identification}
\label{fig:1b}
\end{subfigure}

\vspace{0.5cm}

\begin{subfigure}{0.48\textwidth}
\centering
\includegraphics[height=2in]{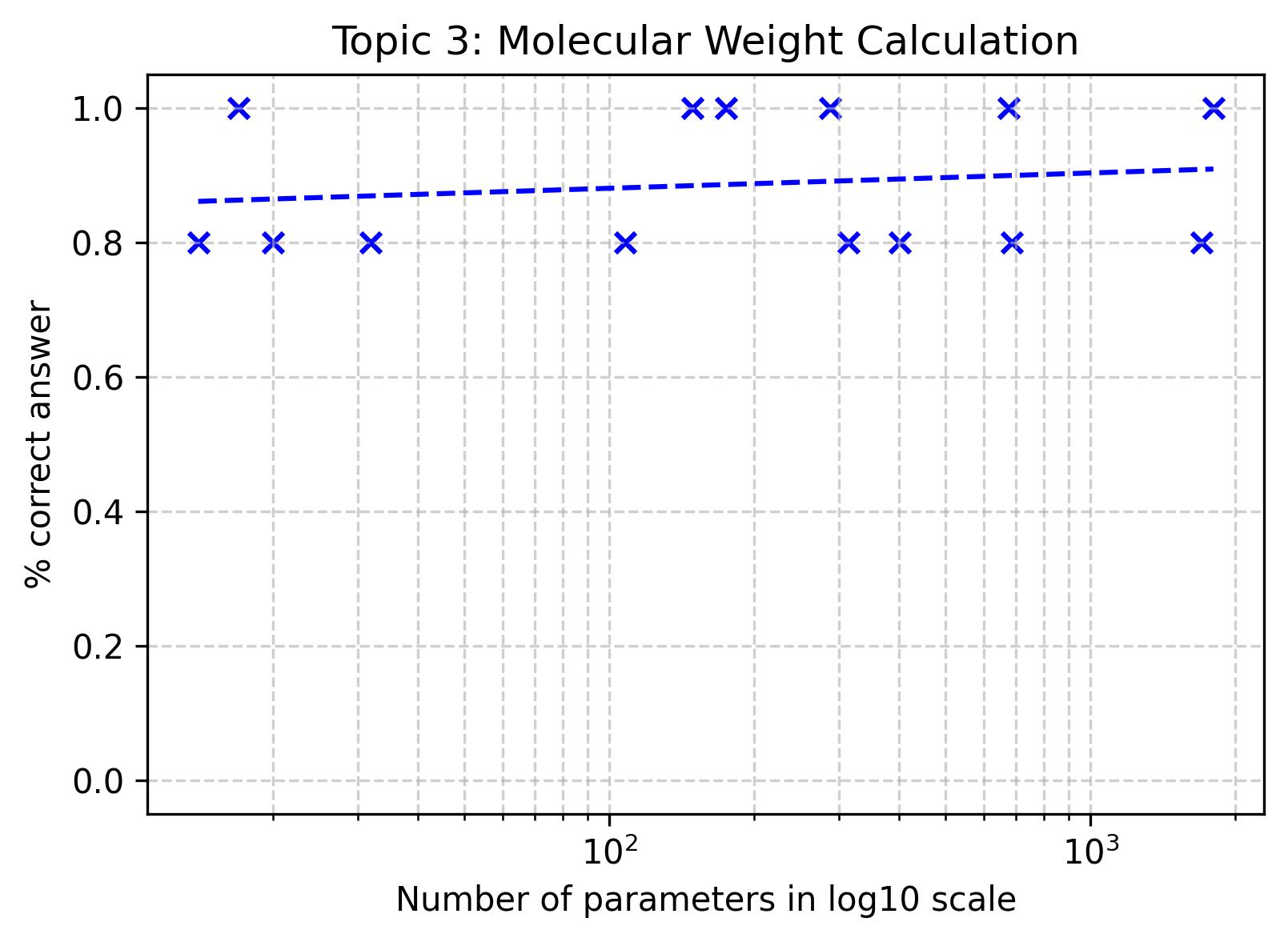}
\caption{Topic 3: Calculating the Molecular Weight}
\label{fig:1c}
\end{subfigure}
\hfill
\begin{subfigure}{0.48\textwidth}
\centering
\includegraphics[height=2in]{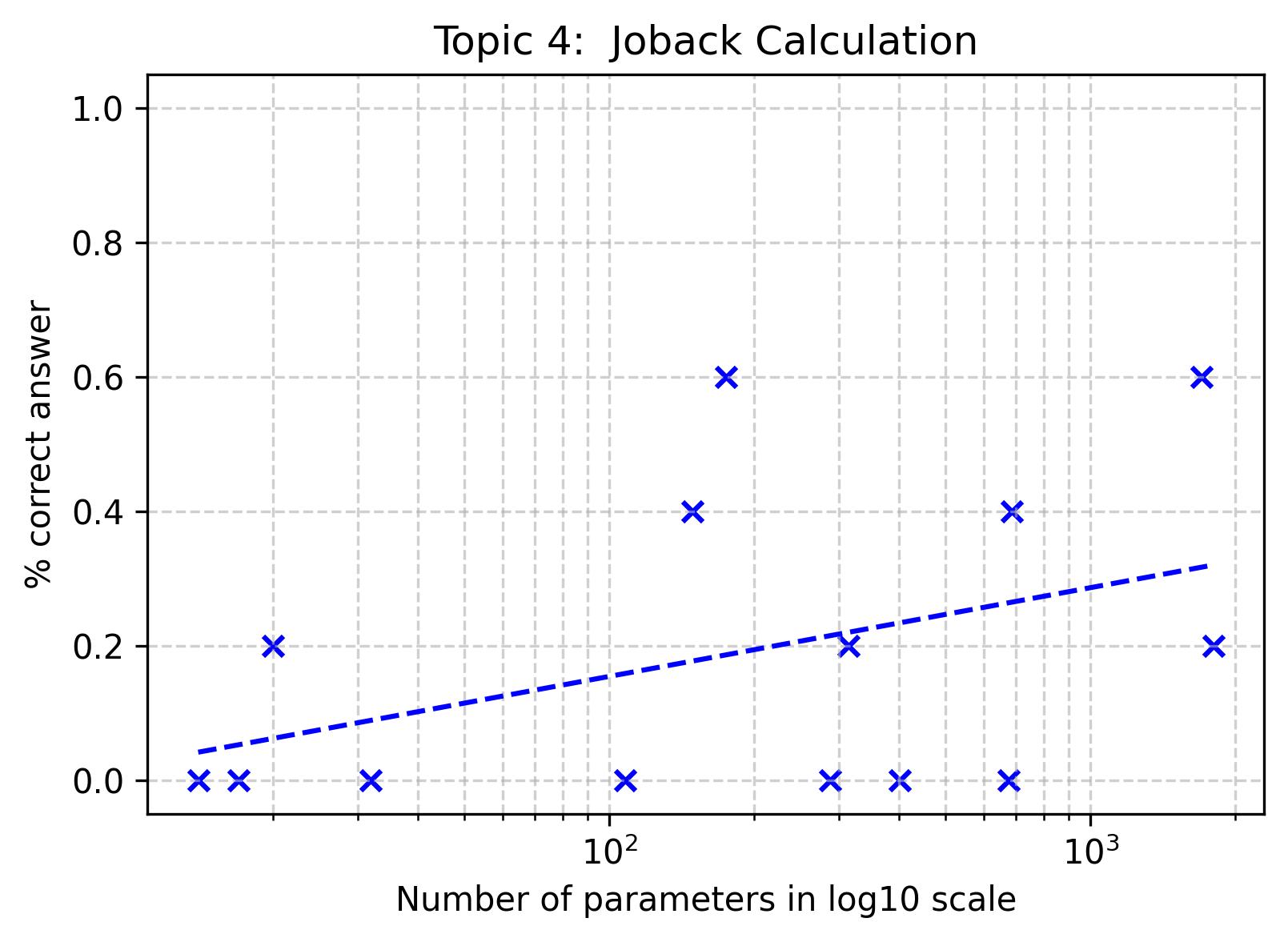}
\caption{Topic 4: Boiling Point by the Joback Method}
\label{fig:1d}
\end{subfigure}

\caption{Relationship between LLM accuracy and model parameter scale.}

\end{figure}

\section{Main Results}

\subsection{Calculating Sigma and Pi Bonds in Cosmetic Ingredients}

Sigma and pi bonds form the foundational framework of molecular structure and play a central role in determining the physical and chemical properties of compounds \citep{Carey:2023aa}. Each LLM was asked: ``\textsl{In one simple sentence, state the number of sigma and pi bonds in \{ingredient\}}''. Five cosmetic ingredients were selected, spanning simple and complex molecular structures to probe how each model handles varying levels of chemical complexity. Bond counts for all five ingredients were independently verified through manual analysis and cross-referenced against NCBI PubChem databases \citep{Kim:2023aa}. The verified ground-truth values are detailed in Appendix ~\ref{tab:bond_counts}.


The relationship between model scale and accuracy is visualized in Figure~\ref{fig:1a}. For this metric, a binary ``all-or-nothing'' scoring system was applied: a LLM scores 1 only if both bond counts were precisely correct. Only five models, the GPT-5 series, the Grok-4 series, and Gemini 2.5 Pro, maintained perfect accuracy across all stimuli.
While the data suggests a general upward trajectory, implying that raw parameter scaling bolsters chemical reasoning, the slope of the regression fell just short of traditional statistical significance ($p = 0.058$). This near-miss is philosophically telling; it suggests that while brute-force scaling improves performance, it does not inherently bridge the "reasoning gaps" found in specialized scientific domains.

The performance of non-U.S.-based models was particularly noteworthy and perhaps a bit sobering. Despite its massive architectural footprint, Mistral Large failed to produce a single correct response ($0\%$ accuracy), while DeepSeek-R1 achieved only $40\%$. This disparity invites a critical look at potential "corpus bias." Although English is the lingua franca of modern science, models trained on European (Mistral) or Chinese (DeepSeek) dominant datasets may utilize tokenization strategies or training weights that do not map seamlessly to English-centric IUPAC nomenclature and bonding conventions \citep{iupac2014bluebook}.  This disparity warrants further investigation into how different development environments shape a model's internalization of Western chemical standards.

\subsection{Identification of Synthetic Preservatives in Cosmetic Formulations}

Ingredient transparency is increasingly important to consumers, particularly those with sensitivities to specific preservatives who need to know precisely what a product contains \citep{Lundov:2009aa}. Each LLM was therefore asked: "\textsl{In one sentence, list all the synthetic preservatives found in \{this product\}.}" Five commercially available products were selected, covering a toner, cleanser, facial cream, foundation, and lotion. Correct responses were verified by cross-referencing each product's official ingredient list with the INCIDecoder database \citep{INCIDecoder:2025aa}, a research-backed resource that classifies cosmetic ingredients by function, safety profile, and irritancy potential. Full product details and verified preservative profiles are provided in Appendix~\ref{tab:preservatives}.

A response was scored as accurate only when all recognized synthetic preservatives in the product were correctly identified. As shown in Figure~\ref{fig:1b}, accuracy rates were generally low across the 14 LLMs, suggesting that most models struggled to reliably identify synthetic preservatives in real cosmetic formulations. 
Regression analysis revealed no significant relationship between parameter count and task performance, with uniform deficits observed across all model scales ($p = 0.34$).
Since web access was disabled via the API, all models had to rely solely on their pre-existing training knowledge \citep{gao2024rag_survey}, which implies a limitation in specialized domains that training data may not comprehensively capture product-specific ingredient information. The low accuracy rates observed here likely reflect gaps in training data coverage rather than a fundamental deficit in reasoning. This interpretation is supported by supplementary testing using web-enabled LLM interfaces. The results were stark: Google Gemini, for example, achieved $100\%$ accuracy when granted real-time internet access. The results suggest that real-time retrieval, rather than parametric knowledge alone, accounts for much of the performance gap.

\subsection{Calculating the Molecular Weight of Cosmetic Ingredients}

Molecular weight is a foundational concept in general chemistry, calculated by summing the atomic weights of all atoms in a molecule \citep{Zumdahl:2018aa}. In cosmetics, it plays a meaningful role in determining how an ingredient interacts with the skin and influences the viscosity and sensory feel of a formulation \citep{Walters:2002aa, Rosen:2015aa}. Each LLM was asked: "\textsl{In one sentence, provide the molecular weight of \{ingredient\}, rounded to one decimal place.}" Five commonly used cosmetic ingredients were selected, including Squalane, Octane, Dodecane, Isopropyl Alcohol, and Chlorphenesin, with their molecular weights independently verified through manual calculation and cross-referencing against the NCBI PubChem database \citep{Kim:2023aa}. The confirmed ground-truth values are provided in Appendix~\ref{tab:molecular_weight}.

This is a relatively straightforward calculation for anyone with a basic chemistry background, and the results largely reflected that. Six out of 14 LLMs achieved 100\% accuracy, and the remaining models correctly identified 4 out of 5 molecular weights (see Figure~\ref{fig:1c}), an encouraging outcome overall, though it does suggest that even simple numerical tasks are not entirely error-free across all models. Notably, parameter count did not predict performance, as both smaller and larger models achieved perfect accuracy ($p=0.58$), pointing to training data quality and architectural design as more meaningful determinants than model scale alone. Performance on this task was considerably better than on the synthetic preservative identification task, reinforcing the broader pattern that LLMs tend to be more reliable for structured calculations grounded in widely available chemical data than for recalling specific product-level information that may be sparsely represented in their training corpora.

\subsection{Calculating the Boiling Point of Cosmetic Ingredients using the Joback Method}

Boiling points have direct practical relevance in cosmetic formulation, influencing product stability, texture, and performance during both manufacturing and skin application \citep{Walters:2002aa, Rosen:2015aa}. To probe whether LLMs can move beyond simple data retrieval into genuine structural reasoning, each model was asked: "\textsl{In one sentence, calculate the normal boiling point of \{ingredient\} in Kelvin using the Joback method and provide the answer as a whole number}." Five ingredients were selected: Squalane, Propylene Glycol, Dodecane, Isopropyl Alcohol, and Chlorphenesin. 

The Joback method was chosen because it is a deterministic group-contribution technique that estimates thermophysical properties solely from a molecule's functional groups and architecture \citep{Joback:1987aa}. Experimentally measured boiling points are unavailable for Squalane and Chlorphenesin due to their high molecular weights or thermal decomposition prior to boiling; the Joback method therefore provides a verifiable ground truth. 
Benchmark values were derived using the DDBST Online Property Estimation tool and cross-verified through manual calculation \citep{DDBST-GmbH:2025aa}; full details are provided in the Appendix~\ref{tab:joback_boiling_points}.

Answering correctly required a multi-step process: decomposing each molecule into its constituent functional groups, retrieving the corresponding Joback contribution values, executing the summation, and applying the appropriate unit conversion and rounding. Despite the prescriptive nature of the task, the LLMs demonstrated a systemic inability to perform these operations. As shown in Figure~\ref{fig:1d}, 50\% of the models (7 out of 14) failed to return a single correct value. Only two models, GPT-5 and Grok-4, attained a peak accuracy of 60\%. As observed in previous tasks, model scale was not a predictor of success ($p=0.14$); larger ``frontier'' models struggled as significantly as their smaller counterparts.

A particularly notable pattern was the tendency of several models to hallucinate Joback group contribution constants, suggesting that while LLMs can often retrieve general chemical facts, they struggle with algorithmic tasks requiring precise structural decomposition and sequential arithmetic. In applied contexts such as thermal safety assessment or formulation stability prediction, these findings underscore the necessity of independent human verification of any LLM-generated thermophysical data.

\subsection{Qualitative Analysis of LLM Responses to Skincare Scenarios}

Beyond quantitative tasks, the study evaluated how well LLMs handle the practical skincare questions that everyday consumers are likely to ask. Two qualitative topics were assessed: sequential skincare advice for individuals with sensitive and acne-prone skin, and first-step recommendations for treating common skin concerns.  These topics were chosen because they reflect frequently encountered consumer questions and offer a meaningful window into whether LLMs can provide guidance consistent with established dermatological principles. While such questions are ideally directed to a dermatologist, specialist access is not always practical, and LLMs have shown growing promise in bridging the gap between consumers and professional skincare guidance \citep{Ferreira:2023aa}.

Since these topics are inherently qualitative, responses were assessed for alignment with broadly accepted skincare principles rather than scored against a single correct answer. Trigram analysis (three-word phrase frequency) was applied across all LLM responses for each topic, with word cloud visualizations used to highlight areas of consensus and divergence \citep{Heimerl:2014aa}.

\begin{figure}[t]
     \centering
     \begin{subfigure}[b]{0.48\textwidth}
         \centering
         \includegraphics[height=2.2in]{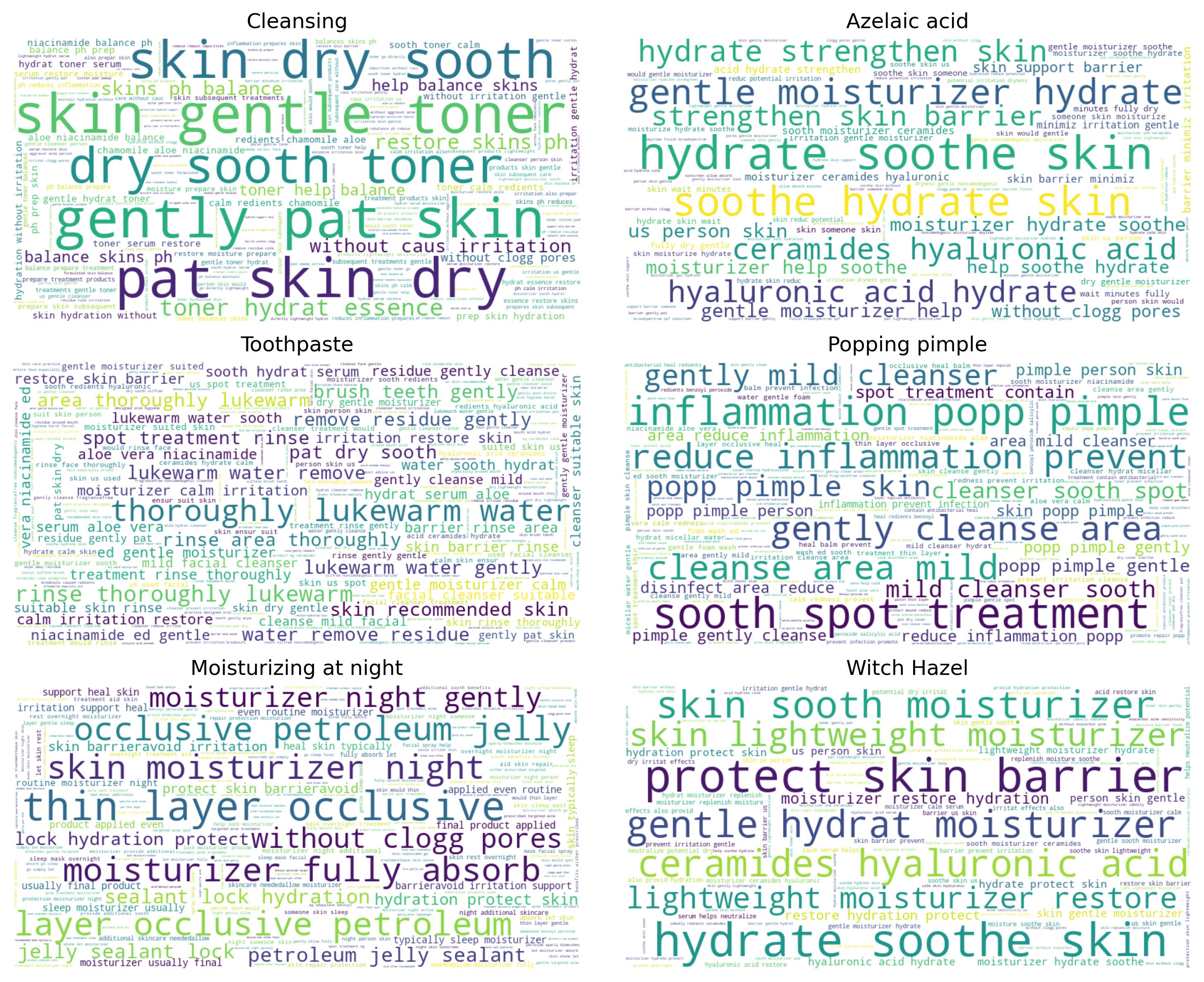}
         \caption{Sequential skincare advice}
         \label{fig:sequential_application}
     \end{subfigure}
     \hfill 
     \begin{subfigure}[b]{0.48\textwidth}
         \centering
         \includegraphics[height=2.2in]{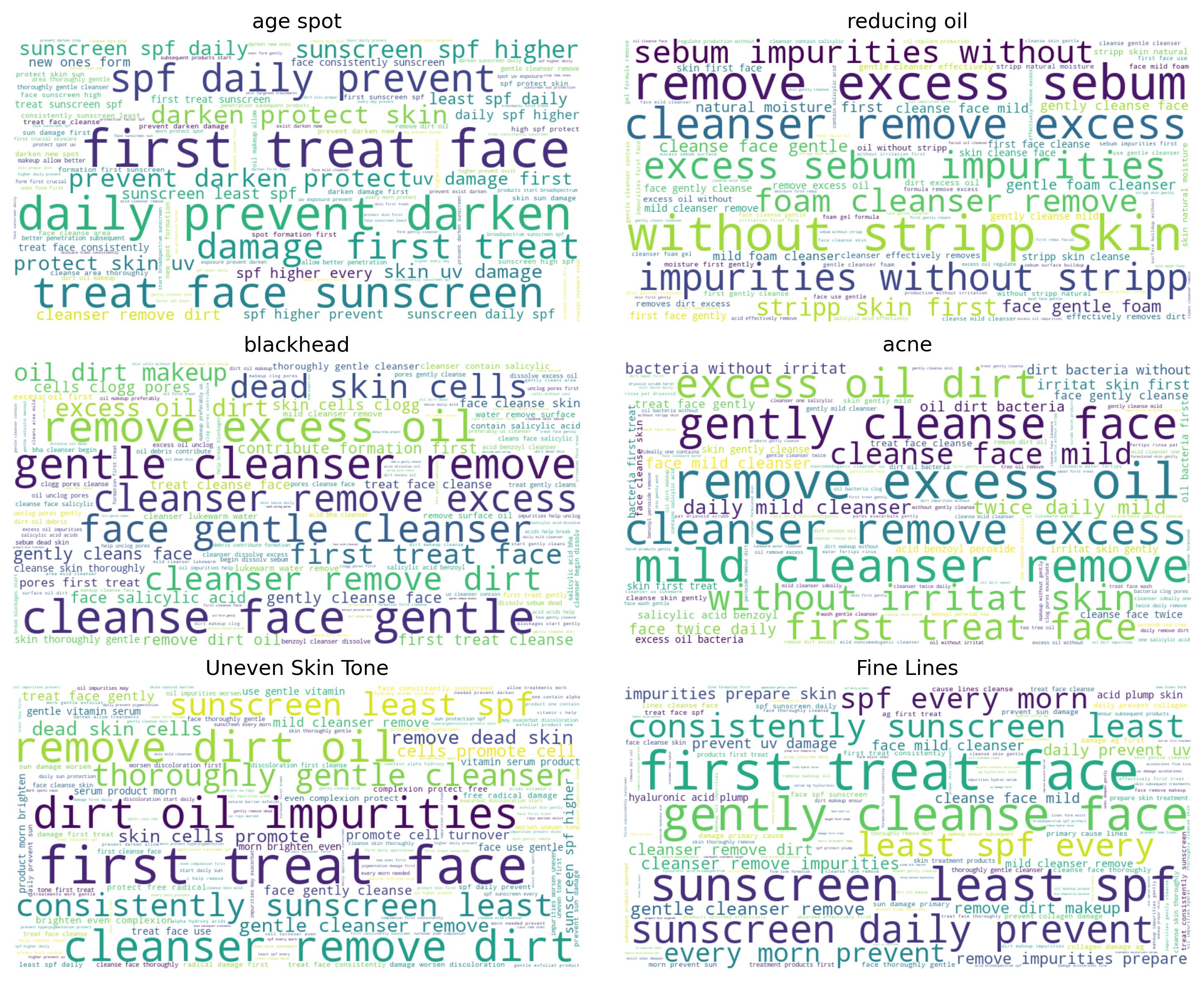}
         \caption{First-step recommendations for skin concerns}
         \label{fig:initial_treatments}
     \end{subfigure}
     
     \caption{N-gram analysis ($N=3$) of qualitative LLM responses on Topic 5 regarding consumer skincare advice. The size of trigrams in the word cloud is proportional to their frequency of occurrence across the 14 LLMs.}
     \label{fig:quantitative_analysis} 
\end{figure}

The results are shown in Figures~\ref{fig:sequential_application} and \ref{fig:initial_treatments}. For the first topic, the recommended next step for sensitive, acne-prone skin, LLMs' responses following cleansing consistently emphasized gentle handling and barrier protection, with common trigrams including ``pat skin dry,'' ``gently pat skin,'' and ``skin gentle toner.'' After azelaic acid application, responses were more varied but centered on hydration, referencing ``gentle moisturizer hydrate,'' ``hydrate soothe skin,'' and ``hyaluronic acid.'' Responses following toothpaste application showed little consensus, though washing with lukewarm water was the most commonly mentioned step. After popping a pimple, common themes included ``reduce inflammation,'' ``gently cleanse area,'' and ``spot treatment.'' Following nighttime moisturizing, the most frequent trigrams referenced occlusive layering, while responses after witch hazel application emphasized barrier support and humectants such as ceramides and hyaluronic acid.

For the second topic, first-step recommendations for common skin concerns, responses were notably more consistent. Sun protection dominated recommendations for both pigmentary changes and anti-aging concerns, with repeated references to daily sunscreen use and the prevention of further darkening. For acne and excess oil, LLMs consistently recommended gentle cleansing without compromising skin integrity.

Notably, most LLMs appended explicit disclaimers recommending consultation with a board-certified dermatologist, a meaningful signal that these models recognize the boundaries of medical guidance. The broad framing of the questions also appeared to naturally discourage prescriptive, personalized recommendations, given the diversity of individual skin types, medical histories, and sensitivities. Taken together, these observations suggest that while LLMs fall short on precise chemical reasoning, they have been meaningfully aligned to operate within a safety-conscious framework when addressing general consumer skincare questions.

\section{Discussion and Conclusions}

\subsection{Discussion}

Can LLMs reliably bridge the information gap between complex cosmetic chemistry and the general public? Our findings suggest that the breadth of their training does not necessarily translate into deep mastery of specialized scientific concepts. A clear and consistent pattern emerged: strong performance on familiar, retrieval-based tasks, and significant failure on those requiring structural reasoning or algorithmic precision.

\textbf{The Disparity Between Retrieval and Reasoning}.  LLMs performed well on deterministic tasks such as calculating molecular weights and identifying sigma and pi bonds — tasks where the answer is a direct lookup from well-represented training data. Performance dropped sharply, however, when the Joback method was required, a technique that demands functional group decomposition, parameter retrieval, and sequential arithmetic. This points to a meaningful architectural gap: these models can retrieve chemical constants with reasonable confidence, but they are not yet reliably capable of applying them through multi-step structural reasoning. The persistent difficulty in identifying synthetic preservatives compounds this concern, as ingredient transparency is increasingly important for consumers navigating sensitivities and health-conscious purchasing decisions 
\citep{Lundov:2009aa}.

By contrast, LLMs performed considerably better on general skincare topics, providing broadly sensible guidance across a range of consumer scenarios. This likely reflects the relative abundance of skincare-related content in their training corpora \citep{Zhao:2023aa}. These responses were, however, general in nature and lacked the clinical specificity needed for personalized dermatological guidance.

\textbf{LLM Commercialization and Consumer Safety}.
Although most LLMs appropriately included disclaimers advising users to seek professional guidance, the confident, authoritative tone these models tend to adopt may still lead consumers to place unwarranted trust in overly broad answers \citep{Ferreira:2023aa}. This reflects a broader concern about the incentive structures shaping LLM development. In a competitive commercial landscape, responsiveness and perceived helpfulness may be prioritized over scientific rigor. \citet{Ouyang:2022aa}
demonstrated that fine-tuning models to follow instructions more agreeably can degrade performance on strictly logical tasks, and \citet{Fanous:2025aa} found that sycophantic behavior, where models adjust responses to align with user expectations, is measurable and widespread across leading LLMs, posing genuine risks in domains where accuracy matters.

\textbf{Hallucination and the Illusion of Confidence}. A recurring pattern across experiments was the tendency of LLMs to generate plausible-sounding but factually incorrect numerical or chemical data. This is particularly concerning in cosmetic and skincare contexts, where consumers often interpret confident responses as reliable ones. While \citet{Kim:2025aa} attribute medical hallucinations primarily to failures in temporal reasoning, the hallucinations observed here appear to have a more foundational cause, namely, gaps in domain-specific training data, as seen with synthetic preservatives, and an inability to execute structured chemical algorithms with precision.

\textbf{The Role of Model Scale and Architecture}.
A notable finding was that model scale did not reliably predict performance. In several instances, smaller ``flash'' or ``mini'' models matched or outperformed their larger counterparts, consistent with the scaling laws proposed by \citet{Hoffmann:2022aa}, who emphasize that training data quality and architectural efficiency are as consequential as raw parameter count. Even with carefully standardized single-turn prompts, performance remained inconsistent across models, pointing to differences in training data composition and optimization objectives as the more meaningful sources of variability.

\subsection{Conclusions and Recommendations}

LLMs have become capable and widely accessible information tools, but they are not yet reliable sources for cosmetic chemistry or precision skincare guidance. The gap between their fluency in general advice and their limitations in rigorous chemical reasoning creates a superficial appearance of expertise that could genuinely mislead consumers. Two practical recommendations follow from these findings.
\begin{itemize}
    \item For AI developers, there is a compelling case for prioritizing scientific grounding over conversational agreeableness. Incorporating specialized chemical datasets, domain-adaptive pre-training, and clearer internal distinctions between evidence-based science and anecdotal content would meaningfully improve reliability in this domain \citep{Cao:2025aa}.

    \item For consumers, LLM-generated information is best treated as a starting point rather than a final answer. For questions involving synthetic preservatives, complex formulations, or clinical skin conditions, the judgment of a qualified cosmetic chemist or dermatologist remains indispensable.

\end{itemize}

Until LLMs can more reliably connect coherent language to rigorous chemical reasoning, their role in cosmetic chemistry should remain supplementary. In a field where precise chemical details carry real implications for consumer health and product safety, accuracy should never be compromised for the sake of a more conversational user experience.

\section*{Acknowledgments}
The author acknowledges the assistance of ChatGPT, Gemini, and Claude in refining English stylistics and enhancing grammar.

\bibliography{llmcosmetic}

@article{gao2024rag_survey,
	author = {Gao, Yunfan and Xiong, Yun and Gao, Xinyu and Jia, Kangxiang and Pan, Jinliu and Bi, Yutao and Dai, Yi and Sun, Jianyuan and Wang, Meng and Wang, Haofen},
	doi = {10.48550/arXiv.2312.10997},
	journal = {arXiv preprint arXiv:2312.10997},
	title = {Retrieval-Augmented Generation for Large Language Models: A Survey},
	url = {https://arxiv.org/abs/2312.10997},
	year = {2024}}

@book{iupac2014bluebook,
	address = {Cambridge},
	author = {Favre, Henri A and Powell, Warren H},
	doi = {10.1039/9781849733069},
	isbn = {978-0-85404-182-4},
	publisher = {Royal Society of Chemistry},
	title = {Nomenclature of Organic Chemistry: IUPAC Recommendations and Preferred Names 2013},
	year = {2014}}

@inproceedings{lewis2020rag,
	author = {Lewis, Patrick and Perez, Ethan and Piktus, Aleksandra and Petroni, Fabio and Karpukhin, Vladimir and Goyal, Naman and K{\"u}ttler, Heinrich and Lewis, Mike and Yih, Wen-tau and Rockt{\"a}schel, Tim and Riedel, Sebastian and Kiela, Douwe},
	booktitle = {Advances in Neural Information Processing Systems},
	pages = {9459--9474},
	title = {Retrieval-Augmented Generation for Knowledge-Intensive {NLP} Tasks},
	url = {https://proceedings.neurips.cc/paper/2020/file/6ad8b6d035414619ca24844391696417-Paper.pdf},
	volume = {33},
	year = {2020}}

@article{goktas2024assessing,
	author = {Goktas, Pinar and Grzybowski, Andrzej},
	doi = {10.3390/jcm13051336},
	journal = {Journal of Clinical Medicine},
	number = {5},
	pages = {1336},
	publisher = {MDPI},
	title = {Assessing the Impact of {ChatGPT} in Dermatology: {A} Comprehensive Rapid Review},
	volume = {13},
	year = {2024}}

@inproceedings{vaswani2017attention,
	author = {Vaswani, Ashish and Shazeer, Noam and Parmar, Niki and Uszkoreit, Jakob and Jones, Llion and Gomez, Aidan N. and Kaiser, {\L}ukasz and Polosukhin, Illia},
	booktitle = {Advances in Neural Information Processing Systems},
	pages = {5998--6008},
	title = {Attention is All You Need},
	url = {https://proceedings.neurips.cc/paper/2017/file/3f5ee243547dee91fbd053c1c4a845aa-Paper.pdf},
	volume = {30},
	year = {2017}}

@article{Naveed:2023aa,
	author = {Naveed, Humza and Khan, Asad Ullah and Qiu, Shi and Saqib, Muhammad and Anwar, Saeed and Usman, Muhammad and Akhtar, Naveed and Barnes, Nick and Mian, Ajmal},
	journal = {arXiv preprint arXiv:2307.06435},
	title = {A Comprehensive Overview of Large Language Models},
	url = {https://doi.org/10.48550/arXiv.2307.06435},
	year = {2023}}

@misc{Hostinger:2025aa,
	author = {{Hostinger}},
	howpublished = {Hostinger International Ltd.},
	title = {{LLM} Statistics: Key Facts and Figures for 2025--2026},
	url = {https://www.hostinger.com/tutorials/llm-statistics},
	year = {2025}}

@misc{Toloka-AI:2025aa,
	author = {{Toloka AI}},
	howpublished = {Toloka Blog},
	title = {History of LLMs: Complete Timeline \& Evolution (1950--2026)},
	url = {https://toloka.ai/blog/history-of-llms/},
	year = {2025}}

@misc{OpenAI:2026aa,
	author = {{OpenAI}},
	month = {February},
	title = {Scaling AI for Everyone},
	url = {https://openai.com/index/scaling-ai-for-everyone/},
	year = {2026}}

@book{Rosen:2015aa,
	editor = {Rosen, Meyer R.},
	publisher = {Chemical Publishing Co.},
	title = {Harry's Cosmeticology, 9th Edition: Focus Books --- Handbook of Skin Anti-aging Theories for Cosmetic Formulation Development},
	volume = {1},
	year = {2015}}

@misc{Carrara:2026aa,
	author = {Carrara, Alessandro},
	howpublished = {Cosmetics Business},
	month = {January},
	title = {{AI} is Strengthening its Grip on Beauty in 2026: Here's Why{\ldots}},
	url = {https://cosmeticsbusiness.com/ai-is-strengthening-its-grip-on-beauty},
	year = {2026}}

@article{Goh:2023aa,
	author = {Goh, C. L. and Wu, Y. and Welsh, B. and Kerrouche, N. and Lu, C. and Wang, X. and Sangueza, M. and Lee, J. B. and See, P. and Kang, H. N.},
	doi = {10.1111/jocd.15519},
	journal = {Journal of Cosmetic Dermatology},
	number = {1},
	pages = {45--54},
	title = {Expert Consensus on Holistic Skin Care Routine: Focus on Acne, Rosacea, Atopic Dermatitis, and Sensitive Skin Syndrome},
	volume = {22},
	year = {2023}}

@article{Leon:2023aa,
	author = {Leon, H. E. and Pimentel, A. S.},
	doi = {10.1021/acs.jcim.3c00285},
	journal = {Journal of Chemical Information and Modeling},
	number = {6},
	pages = {1649--1655},
	title = {Do Large Language Models Understand Chemistry? A Conversation with ChatGPT},
	volume = {63},
	year = {2023}}

@inproceedings{Guo:2023aa,
	author = {Guo, Taichen and Guo, Kehan and Nan, Bozhao and Liang, Zhenwen and Guo, Zhichun and Chawla, Nitesh V. and Wiest, Olaf and Zhang, Xiangliang},
	booktitle = {Advances in Neural Information Processing Systems},
	pages = {59662--59688},
	title = {What Can Large Language Models Do in Chemistry? A Comprehensive Benchmark on Eight Tasks},
	url = {https://doi.org/10.48550/arXiv.2305.18365},
	volume = {36},
	year = {2023}}

@article{Ferreira:2023aa,
	author = {Ferreira, A. L. and Chu, B. and Grant-Kels, J. M. and Ogunleye, T. and Lipoff, J. B.},
	doi = {10.2196/49280},
	journal = {JMIR Dermatology},
	pages = {e49280},
	title = {Evaluation of ChatGPT Dermatology Responses to Common Patient Queries},
	volume = {6},
	year = {2023}}

@article{Hoffmann:2022aa,
	author = {Hoffmann, Jordan and Borgeaud, Sebastian and Mensch, Arthur and Buchatskaya, Elena and Cai, Trevor and Rutherford, Eliza and de Las Casas, Diego and Hendricks, Lisa Anne and Welbl, Johannes and Clark, Aidan and others},
	journal = {arXiv preprint arXiv:2203.15556},
	title = {Training Compute-Optimal Large Language Models},
	url = {https://arxiv.org/abs/2203.15556},
	year = {2022}}

@article{Kim:2023aa,
	author = {Kim, Sunghwan and Chen, Jie and Cheng, Tiejun and Gindulyte, Asta and He, Jia and He, Siqian and Li, Qingliang and Shoemaker, Shoemaker and others},
	doi = {10.1093/nar/gkac956},
	journal = {Nucleic Acids Research},
	number = {D1},
	pages = {D1373--D1380},
	title = {PubChem 2023 Update},
	volume = {51},
	year = {2023}}

@manual{OpenRouter:2025aa,
	author = {{OpenRouter}},
	title = {OpenRouter: A Unified Interface for LLMs [API Platform Documentation]},
	url = {https://openrouter.ai/docs},
	year = {2025}}

@manual{OpenAI:2025ab,
	author = {{OpenAI}},
	title = {OpenAI Python Library [Software]},
	url = {https://pypi.org/project/openai/},
	year = {2025}}

@article{Huang:2024aa,
	author = {Huang, Lei and Yu, Weitao and Ma, Weijia and Zhong, Weihua and Feng, Zhangyin and Wang, Haonan and Chen, Qiang and Peng, Weihua and Feng, Xiaocheng and Qin, Bing and Liu, Ting},
	doi = {10.1145/3703155},
	journal = {ACM Transactions on Information Systems},
	number = {1},
	title = {A Survey on Hallucination in Large Language Models: Principles, Taxonomy, Challenges, and Open Questions},
	volume = {1},
	year = {2024}}

@inproceedings{Heimerl:2014aa,
	author = {Heimerl, Florian and Lohmann, Steffen and Lange, Simon and Ertl, Thomas},
	booktitle = {Proceedings of the 47th Hawaii International Conference on System Sciences (HICSS)},
	doi = {10.1109/HICSS.2014.231},
	pages = {1833--1842},
	title = {Word Cloud Explorer: Text Analytics Based on Word Clouds},
	year = {2014}}

@book{Carey:2023aa,
	author = {Carey, Francis A. and Giuliano, Robert M. and Allison, Neil T. and Bane, Susan L.},
	edition = {12th},
	publisher = {McGraw Hill},
	title = {Organic Chemistry},
	year = {2023}}

@misc{INCIDecoder:2025aa,
	author = {{INCIDecoder}},
	title = {INCIDecoder: Decode Your Skincare Ingredients [Online Cosmetic Ingredient Database]},
	url = {https://incidecoder.com},
	year = {2025}}

@article{Joback:1987aa,
	author = {Joback, Kevin G. and Reid, Robert C.},
	doi = {10.1080/00986448708960487},
	journal = {Chemical Engineering Communications},
	number = {1--6},
	pages = {233--243},
	title = {Estimation of Pure-Component Properties from Group-Contributions},
	volume = {57},
	year = {1987}}

@book{Zumdahl:2018aa,
	author = {Zumdahl, Steven S. and Zumdahl, Susan A. and DeCoste, Donald J.},
	edition = {10th},
	publisher = {Cengage Learning},
	title = {Chemistry},
	year = {2018}}

@incollection{Walters:2002aa,
	author = {Walters, Kenneth A. and Roberts, Michael S.},
	booktitle = {Dermatological and Transdermal Formulations},
	editor = {Walters, Kenneth A.},
	pages = {1--40},
	publisher = {Marcel Dekker},
	title = {The Structure and Function of Skin},
	year = {2002}}

@misc{DDBST-GmbH:2025aa,
	author = {{DDBST GmbH}},
	howpublished = {Dortmund Data Bank Software \& Separation Technology},
	title = {Online Property Estimation --- Normal Boiling Point [Computational Tool]},
	url = {http://ddbonline.ddbst.de/OnlinePropertyEstimation/OnlinePropertyEstimation.exe},
	year = {2025}}

@article{Lundov:2009aa,
	author = {Lundov, M. D. and Moesby, L. and Zachariae, C. and Johansen, J. D.},
	doi = {10.1111/j.1600-0536.2008.01501.x},
	journal = {Contact Dermatitis},
	number = {2},
	pages = {70--78},
	title = {Contamination Versus Preservation of Cosmetics: A Review on Legislation, Usage, Infections, and Contact Allergy},
	volume = {60},
	year = {2009}}

@article{Zhao:2023aa,
	author = {Zhao, Wayne Xin and Zhou, Kun and Li, Junyi and Tang, Tianyi and Wang, Xiaolei and Hou, Yupeng and Min, Yingqian and Zhang, Beichen and Zhang, Junjie and Dong, Zican and others},
	journal = {arXiv preprint arXiv:2303.18223},
	title = {A Survey of Large Language Models},
	url = {https://arxiv.org/abs/2303.18223},
	year = {2023}}

@inproceedings{Ouyang:2022aa,
	author = {Ouyang, Long and Wu, Jeffrey and Jiang, Xu and Almeida, Diogo and Wainwright, Carroll L. and Mishkin, Pamela and Zhang, Chong and Agarwal, Sandhini and Slama, Katarina and Ray, Alex and others},
	booktitle = {Advances in Neural Information Processing Systems},
	pages = {27730--27744},
	title = {Training Language Models to Follow Instructions with Human Feedback},
	url = {https://arxiv.org/abs/2203.02155},
	volume = {35},
	year = {2022}}

@inproceedings{Fanous:2025aa,
	author = {Fanous, A. and Goldberg, J. and Agarwal, A. A. and Lin, J. and Zhou, A. and Daneshjou, R. and Koyejo, S.},
	booktitle = {Proceedings of the AAAI/ACM Conference on AI, Ethics, and Society (AIES '25)},
	doi = {10.1145/3735632},
	title = {SycEval: Evaluating LLM Sycophancy},
	year = {2025}}

@article{Kim:2025aa,
	author = {Kim, Y. and Jeong, H. and Chen, S. and Li, S. S. and Lu, M. and Alhamoud, K. and Mun, J. and Grau, C. and Jung, M. and Gameiro, R. and others},
	doi = {10.1101/2025.02.28.25323115},
	journal = {medRxiv},
	title = {Medical Hallucinations in Foundation Models and Their Impact on Healthcare},
	year = {2025}}

@article{Cao:2025aa,
	author = {Cao, S. and Shi, W. and Yu, Q. and Li, Z. and Zhao, H. and Cai, H. and Wen, J.-R.},
	journal = {ACS Omega},
	title = {Leveraging Prompt Engineering in Large Language Models for Accelerating Chemical Research},
	url = {https://pmc.ncbi.nlm.nih.gov/articles/PMC12022906/},
	year = {2025}}
\bibliographystyle{colm2026_conference}

\appendix

\section{Appendix: Experimental Prompts and Target Stimuli}
\label{appendix:prompts}

\subsection{Quantitative Inquiries}
\begin{itemize}
    \item \textbf{Bond Analysis:} ``In one simple sentence, state the number of sigma and pi bonds in [Ingredient].''
    \item \textbf{Preservative Identification:} ``In one sentence, list all the synthetic preservatives found in [Product Name].''
    \item \textbf{Molecular Weight:} ``In one sentence, provide the molecular weight of [Ingredient], rounded to one decimal place.''
    \item \textbf{Joback Calculation:} ``In one sentence, calculate the normal boiling point of [Ingredient] in Kelvin using the Joback method and provide the answer as a whole number.''
\end{itemize}

\subsection{Qualitative Scenarios}
\begin{itemize}
    \item \textbf{Sequential skincare advice:} ``In one sentence, briefly describe the immediate next skincare step after using/applying [Scenario] for a person with sensitive and acne-prone skin.''
    \item \textbf{First-step recommendations:} ``In one sentence, briefly describe the immediate first skincare step for treating [Scenario] on the face.''
\end{itemize}

\section{Appendix: Correct Answers for Quantitative Inquiries}
\label{appendix:correct_answers}

\subsection{Calculating Sigma and Pi Bonds}
    \label{tab:bond_counts}

\begin{table}[H]
    \centering
    \caption{Manual Calculation of sigma and pi Bond Counts for Selected Cosmetic Ingredients.}
    \begin{tabular}{llcc}
        \toprule
        \textbf{Ingredient} & \textbf{Molecular Formula} & \textbf{Sigma Bonds} & \textbf{Pi Bonds} \\ 
        \midrule
        Squalane              & $C_{30}H_{62}$             & 90 & 0 \\
        Phenoxyethanol        & $C_{8}H_{10}O_{2}$         & 20 & 3 \\
        Sodium Lauryl Sulfate & $C_{12}H_{25}NaO_{4}S$     & 41 & 2 \\
        Isododecane           & $C_{12}H_{26}$             & 37 & 0 \\
        Chlorphenesin         & $C_{9}H_{11}ClO_{3}$       & 24 & 3 \\ 
        \bottomrule
    \end{tabular}
    \vspace{2pt}
    \begin{flushleft}
    \footnotesize \textit{Note:} Values were independently verified via structural decomposition and cross-referenced with NCBI chemical databases.
    \end{flushleft}
\end{table}

\subsection{Identification of Synthetic Preservatives}
    \label{tab:preservatives}

\begin{table}[H]
    \centering
    \caption{Identification of Synthetic Preservative Systems in Selected Commercial Formulations.}
    \small 
    \begin{tabularx}{\textwidth}{XX}
        \toprule
        \textbf{Cosmetic Product} & \textbf{Verified Synthetic Preservative Ingredients} \\ 
        \midrule
        CeraVe Hydrating Toner & Phenoxyethanol, Ethylhexylglycerin, Chlorophenesin \\
        \addlinespace[0.5em]
        CeraVe Hydrating Cleanser & Phenoxyethanol, Ethylhexylglycerin \\
        \addlinespace[0.5em]
        Kiehl's Ultra Facial Cream & Phenoxyethanol, Ethylhexylglycerin, Chlorophenesin, Salicylic acid \\
        \addlinespace[0.5em]
        Estée Lauder Double Wear & Phenoxyethanol, Sodium dehydroacetate, Pentaerythrityl tetra-di-t-butyl hydroxyhydrocinnamate \\
        \addlinespace[0.5em]
        The Ordinary Hyaluronic Acid 2\% + B5 (2024 reformulation) & Phenoxyethanol, Ethylhexylglycerin, Chlorophenesin, Caprylyl Glycol \\
        \bottomrule
    \end{tabularx}
    \vspace{2pt}
    \begin{flushleft}
    \footnotesize \textit{Note:} Ingredient profiles were cross-referenced against official manufacturer disclosures and the INCIDecoder database. 
    \end{flushleft}
\end{table}

\subsection{Calculating Molecular Weight}
    \label{tab:molecular_weight}

\begin{table}[H]
    \centering
    \caption{Molecular Weight Comparison of Selected Cosmetic Ingredients.}
    \begin{tabular}{llc}
        \toprule
        \textbf{Ingredient} & \textbf{Molecular Formula} & \textbf{Molecular Weight (g/mol)} \\ 
        \midrule
        Squalane          & $C_{30}H_{62}$     & 422.8 \\
        Octane            & $C_{8}H_{18}$      & 114.2 \\
        Dodecane          & $C_{12}H_{26}$     & 170.3 \\
        Isopropyl Alcohol & $C_{3}H_{8}O$      & 60.1  \\
        Chlorphenesin     & $C_{9}H_{11}ClO_{3}$ & 202.6 \\ 
        \bottomrule
    \end{tabular}
    \vspace{2pt}
    \begin{flushleft}
    \footnotesize \textit{Note:} Values were independently verified through stoichiometric summation of standard atomic weights and cross-referenced against the PubChem database.
    \end{flushleft}
\end{table}

\subsection{Calculating Normal Boiling Point using the Joback Method}
    \label{tab:joback_boiling_points}

\begin{table}[H]
    \centering
    \caption{Estimated Normal Boiling Points of Cosmetic Ingredients via the Joback Method.}
    \small 
    \begin{tabularx}{\textwidth}{lllrX}
        \toprule
        \textbf{Ingredient} & \textbf{CAS No.} & \textbf{Formula} & \textbf{$T_b$ (K)*} & \textbf{Common Cosmetic Use} \\ 
        \midrule
        Squalane & 111-01-3 & $C_{30}H_{62}$ & 883 & Moisturizer and anti-aging agent in serums and body oils. \\
        \addlinespace[0.5em]
        Propylene Glycol & 57-55-6 & $C_{3}H_{8}O_{2}$ & 452 & Humectant and solvent used in lotions and hair products. \\
        \addlinespace[0.5em]
        Dodecane & 112-40-3 & $C_{12}H_{26}$ & 474 & Emollient and solvent in skincare and makeup formulations. \\
        \addlinespace[0.5em]
        Isopropyl Alcohol & 67-63-0 & $C_{3}H_{8}O$ & 360 & Astringent and antimicrobial agent in toners and acne treatments. \\
        \addlinespace[0.5em]
        Chlorphenesin & 104-29-0 & $C_{9}H_{11}ClO_{3}$ & 681 & Synthetic antimicrobial preservative used in creams and makeup. \\
        \bottomrule
    \end{tabularx}
    \vspace{2pt}
    \begin{flushleft}
    \footnotesize \textit{Note:} *Boiling points ($T_b$) calculated using the Joback method: $T_b = 198 + \sum \Delta T_{b,i}$. Values rounded to the nearest whole number.
    \end{flushleft}
\end{table}

\end{document}